\documentclass[a4paper]{cas-sc}

\usepackage[numbers]{natbib}
\usepackage{xcolor}
\usepackage{multirow}
\usepackage[normalem]{ulem}
\usepackage{graphicx}
\usepackage{amsmath,amssymb}
\usepackage{booktabs}
\usepackage{hyperref}
\usepackage{tabularx}
\usepackage{array}
\usepackage{subcaption}
\usepackage{float}
\usepackage{capt-of}
\usepackage{algorithm}
\usepackage{algpseudocode}
\usepackage{seqsplit}
\hypersetup{
    colorlinks=true,
    linkcolor=blue,
    citecolor=blue,
    urlcolor=cyan
}

\begin{document}

% Numeric citations are intentionally hyperlinked and coloured blue.
\setcitestyle{numbers,square,comma,sort&compress}
\let\WriteBookmarks\relax

% ----------------------------------------------------------
% Float control for single-column CAS layout
% ----------------------------------------------------------
\renewcommand{\topfraction}{0.85}
\renewcommand{\bottomfraction}{0.65}
\renewcommand{\textfraction}{0.12}
\renewcommand{\floatpagefraction}{0.75}

\setcounter{topnumber}{2}
\setcounter{bottomnumber}{1}
\setcounter{totalnumber}{3}

\shorttitle{DiffusionRef-IMU}
\shortauthors{M.\,M. Golam Hafiz et al.}

\title[mode=title]{DiffusionRef-IMU: Guide-Anchored Residual Diffusion for Cross-Location Inertial Measurement Unit Translation}

\author[1]{M.\,M. Golam Hafiz}[orcid=0009-0003-2795-1033]
\ead{rafihafiz2016@gmail.com}

\author[1]{Safia Akter Simi}
\ead{safiasimi17@gmail.com}

\author[2]{Md. Rakibul Islam}
\cormark[1]
\ead{md.rakibul.islam@mdu.se}

\author[1]{M.\,F. Mridha}[orcid=0000-0001-5738-1631]
\ead{firoz.mridha@aiub.edu}

\affiliation[1]{
    organization={Department of Computer Science,
    American International University-Bangladesh (AIUB)},
    city={Dhaka},
    country={Bangladesh}
}

\affiliation[2]{
    organization={Department of Computer Science and Engineering,
    Mälardalen University},
    city={Västerås},
    state={Västmanland},
    country={Sweden}
}

\cortext[cor1]{Corresponding author: Md. Rakibul Islam
(\href{mailto:md.rakibul.islam@mdu.se}{md.rakibul.islam@mdu.se}).}

\begin{abstract}
Cross-location variation in wearable inertial measurement unit (IMU) signals complicates learning when measurements from a desired target body location are unavailable. This study proposes DiffusionRef-IMU, a two-stage cross-location translation framework in which a deterministic Direct network first estimates the target signal and a guide-anchored residual diffusion model refines only the remaining target discrepancy. The evaluation comprises five subject-disjoint activity--location composition-held-out folds, capacity-matched deterministic and flow-based residual baselines, source-only activity conditioning, computational ablation, and true zero-shot evaluation on two independent datasets. DiffusionRef-IMU with 16-step deterministic DDIM achieved a mean nRMSE of $0.087538\pm0.004922$, compared with 0.090262 for Direct, 0.088847 for deterministic residual refinement, and 0.088417 for eight-step CompFlow, and outperformed each baseline in all five folds. Four-step DDIM retained a mean nRMSE of 0.087820 at 47.62~ms per window, providing a favourable quality--computation trade-off. Replacing ground-truth activity conditioning with a source-only activity predictor (macro-F1 0.8473) increased 16-step DiffusionRef nRMSE only to 0.088190. True zero-shot evaluation produced mixed results: DiffusionRef was marginally favourable on Grouvel2025 but remained worse than Direct on Salda\~na2025. Overall, guide-anchored residual diffusion improves controlled cross-location reconstruction while revealing clear limits in zero-shot external transfer.
\end{abstract}


\maketitle

% ==================================================================
\section{Introduction}\label{sec:introduction}
% ==================================================================
Wearable inertial measurement units (IMUs) provide compact, low-cost measurements of human motion and underpin a wide range of human activity recognition (HAR) systems \cite{napoli2024benchmark}. However, body location can substantially alter the sensed distribution, and transferring a classifier from one anatomical site to another may require additional labelled data or explicit domain adaptation \cite{hegde2023activity}. Recent placement-adaptive work has therefore sought to relax fixed-location assumptions. IMUCoCo, for example, maps flexibly placed IMUs into a shared spatially conditioned representation \cite{zhou2025imucoco}, while W2W takes a complementary simulation-based approach by evaluating IMU placement utility over hundreds of anatomically distributed body-surface locations \cite{ray2025w2w}. These advances underscore the importance of placement variability, but they do not resolve the practical case in which the desired target-location signal is unavailable during training or deployment.

Virtual sensing provides a complementary route by synthesising inertial observations that would otherwise require additional physical instrumentation. WIMUSim models body dynamics, sensor placement and hardware characteristics to generate wearable IMU signals \cite{oishi2025wimusim}. Cross-modality pipelines have also converted text-conditioned human motion into virtual on-body accelerometry for HAR augmentation \cite{leng2023generating}. Their utility, however, depends on the motion regime and on how faithfully local sensor dynamics are reproduced; fine-grained activities can expose limitations that are less apparent for coarser motion \cite{leng2023utility}. Missing-sensor recovery offers a more directly conditioned alternative, with SensorGAN reconstructing unavailable wearable streams from concurrently observed sensors \cite{hussein2024sensorgan}. SynHAR uses personalised biomechanical and human-surface models to synthesise inertial data under placement variation \cite{uhlenberg2024synhar}, while TheraGAN applies conditional adversarial generation to six-channel accelerometer--gyroscope signals for HAR augmentation in therapeutic activities \cite{mohammadzadeh2025theragan}. Cross-location translation is more restrictive: a source-to-target model must recover location-dependent amplitude, orientation, phase and local temporal dynamics while preserving activity-discriminative content. Deterministic regression can collapse uncertain mappings towards central estimates, whereas unconstrained generative synthesis can introduce plausible but target-inaccurate variation. Waveform fidelity and downstream task utility should therefore be assessed separately. Collectively, prior work establishes the value of virtual IMU generation, but augmentation, missing-sensor recovery and placement-invariant classification are not equivalent to reconstructing the specific target-location waveform associated with a specific source-location waveform.

Data-driven generative modelling expands the available design space. Diffusion models have been applied to multivariate inertial generation and, more broadly, to time-series synthesis, forecasting and imputation \cite{oppel2025diffusion,lin2024diffusion}. Flow Matching provides an alternative continuous-transport formulation \cite{lipman2023flow,tong2024improving}. Existing wearable-generation studies, however, primarily address augmentation, imputation or unconditional synthesis rather than paired reconstruction of a specific target body location from a simultaneously observed source location. This distinction raises a further methodological question: whether a generative refiner contributes beyond both a strong deterministic estimate and an equally expressive deterministic residual model.

DiffusionRef-IMU addresses this problem through guide-anchored residual diffusion. A deterministic Direct network first estimates the target-location signal, after which the proposed diffusion model operates only on the residual between this guide and the paired target. The raw source and Direct estimate remain available as persistent conditioning throughout refinement, and deterministic DDIM inference produces the primary reconstruction. Evaluation combines five subject-disjoint composition-held-out folds, deterministic and flow-based residual controls, source-only activity conditioning, inference-cost analysis, and true zero-shot external transfer.

The contributions are as follows:
\begin{itemize}
  \item A guide-anchored residual diffusion formulation for paired cross-location IMU translation in which diffusion models only the correction unresolved by a strong deterministic target estimate.
  \item A controlled five-fold activity--location composition-held-out evaluation showing that DiffusionRef-IMU improves over Direct, capacity-matched deterministic residual refinement, and conditional flow refinement in all five folds.
  \item A quality--computation analysis over 4, 8, 16 and 32 DDIM steps, together with source-only activity conditioning that removes the requirement for ground-truth activity labels at inference.
  \item True zero-shot evaluation on Grouvel2025 and Salda\~na2025 without external training, fine-tuning or normalisation fitting, reporting both favourable and adverse external evidence.
\end{itemize}

The remainder of this paper is organised as follows: Section~\ref{sec:related} reviews related work, Section~\ref{sec:method} presents DiffusionRef-IMU and the evaluation protocol, Section~\ref{sec:results} reports the results, Section~\ref{sec:discussion} discusses the findings and limitations, and Section~\ref{sec:conclusion} concludes the study.

% ==================================================================
\section{Related Work}\label{sec:related}
% ==================================================================
\subsection{Sensor-Placement Variability, Cross-Location Adaptation, and Virtual Sensing}
Wearable-sensor HAR remains sensitive to how and where inertial measurements are acquired. Broad surveys continue to identify variability in users, devices, activities and sensing configurations as barriers to reliable real-world generalisation \cite{haresamudram2025past}, while comparative evaluations show that model choice alone does not remove dependence on sensor configuration and placement \cite{ciortuz2025machine}. Cross-dataset benchmarking likewise demonstrates substantial domain shifts when acquisition conditions change, even when the downstream label space remains similar \cite{napoli2024benchmark}. These observations motivate methods that address placement and domain variation explicitly rather than treating them as incidental noise.

One line of work adapts representations or classifiers across sensing domains. Hegde et al. transfer an activity model from a labelled source body location to an unlabelled target location by aligning source and target embeddings \cite{hegde2023activity}. IMUCoCo instead maps signals from flexibly placed IMUs into a spatially conditioned shared representation for pose estimation and HAR \cite{zhou2025imucoco}, whereas W2W uses large-scale simulation to assess candidate placements across the body surface \cite{ray2025w2w}. More broadly, the transfer-learning literature organises wearable HAR methods around the reuse of information across users, modalities, tasks and domains \cite{dhekane2025transfer}. DWLR addresses wearable domain adaptation under simultaneous feature and label shift through learnable reweighting and separate time- and frequency-domain alignment \cite{li2024dwlr}. COA-HAR combines contrastive self-supervision, sensor-data augmentation and pseudo-label refinement for online test-time adaptation of wearable HAR models \cite{rey2026coahar}. Li et al. address cross-person HAR through deviation-graph constrained non-negative matrix factorisation, using a deviation constraint to mitigate noise and outliers and a local graph to guide transfer between domains \cite{li2025dgnmf}. These approaches improve placement or domain robustness, but their primary objective is activity recognition or placement analysis rather than reconstruction of a target-location waveform.

Other representation-learning strategies target distinct sources of HAR generalisation error. Natural-language supervision has been explored as cross-modal guidance, although sensor heterogeneity and limited descriptive text can constrain its effectiveness relative to supervised or self-supervised alternatives \cite{haresamudram2025limitations}. Rotation-aware Siamese masked reconstruction instead exploits the known transformation of inertial axes to encourage motion representations that are less dependent on sensor-frame orientation \cite{calatrava2026siamese}. These methods show that invariance can be learned without synthesising what another body-mounted sensor would have measured. This distinction is central to DiffusionRef-IMU: classification adaptation asks whether an activity decision transfers, whereas cross-location translation asks whether the time-resolved target signal itself can be reconstructed.

Sensor configuration provides another route to reducing sensing burden. CoSS uses trainable importance scores to jointly select sensor modalities and sampling rates while balancing HAR performance against input and model complexity \cite{liu2026coss}. Its objective is efficient acquisition rather than target-location waveform reconstruction. Virtual sensing addresses unavailable inertial measurements more directly. WIMUSim models body dynamics, placement and sensor characteristics to generate wearable IMU measurements \cite{oishi2025wimusim}, and SensorGAN reconstructs missing wearable streams from concurrently observed sensors \cite{hussein2024sensorgan}. These approaches demonstrate that unavailable sensing information can be approximated, but simulation and multi-sensor recovery are not equivalent to one-source-to-one-target translation under a held-out activity--location composition. The latter requires inference of location-specific amplitude, orientation and temporal structure from a specified observed source while preserving the semantics of the same activity window.

\subsection{Generative Time-Series and Inertial Signal Translation}
Synthetic inertial data can be generated from several starting points. Text-to-motion pipelines convert natural-language activity descriptions into virtual human motion and then into on-body acceleration \cite{leng2023generating}; follow-up evaluation shows that their usefulness depends partly on activity granularity and the difficulty of reproducing subtle local motion \cite{leng2023utility}. SynHAR uses personalised biomechanical dynamics and human-surface models to synthesise inertial signals under placement variation \cite{uhlenberg2024synhar}. MixHAR interpolates labelled and unlabelled wearable behaviours in a calibrated semi-supervised representation space \cite{duan2024wearable}. Simulated multimodal representations have also been combined with mixture-of-experts classification to exploit complementary synthetic cues when labelled wearable data are limited \cite{ray2025simulated}. These methods broaden data coverage, but generation from text, motion models or auxiliary modalities does not enforce agreement with the paired target waveform corresponding to a particular observed source location.

A second family synthesises time series primarily for learning utility. TheraGAN generates multi-axis accelerometer and gyroscope signals for therapeutic activities using a conditional generative adversarial network (GAN), with evaluation covering signal similarity, diversity and downstream HAR augmentation \cite{mohammadzadeh2025theragan}. Hydra-TS uses a multi-agent GAN and spectral representation to optimise multiple synthetic-data objectives, including downstream classification utility \cite{desmet2025hydra}. Diffusion-based inertial generation follows a different mechanism: Oppel and Munz adapt denoising diffusion to multivariate IMU sequences and evaluate the resulting samples through qualitative structure and HAR augmentation \cite{oppel2025diffusion}. A broader survey of diffusion models for time-series applications documents their use in generation, forecasting and imputation, with conditioning and evaluation varying substantially by task \cite{lin2024diffusion}. Improved HAR after augmentation is therefore evidence of task utility, not evidence that a generated sequence reproduces the correct counterfactual measurement at a specified target body site.

Time-series imputation further clarifies this taxonomy. Self-attention-based diffusion has been used to fill partial-blackout patterns by modelling temporal and cross-feature dependencies \cite{islam2025selfattention}, while boundary-enhanced diffusion imputation combines weighted injection with a multi-scale S4-based U-Net to improve continuity and long-range dependency modelling around missing regions \cite{xiao2025boundary}. SensorGAN similarly reconstructs unavailable wearable streams \cite{hussein2024sensorgan}. Imputation is relevant because it estimates unobserved values from observed context, but its missingness pattern is typically defined over channels, features or time spans rather than a body-location translation relation. Cross-location IMU translation therefore lies between deterministic conditional regression and generative reconstruction: the target is paired, yet the mapping can remain ambiguous because local body dynamics differ across locations.

This distinction motivates the deterministic--generative design trade-off. A conditional regressor provides a stable point estimate tied to the observed source, whereas a flexible generator can represent a richer transformation but may also introduce unnecessary deviation from the paired target. Neither behaviour is universally preferable. For reconstruction, a generative mechanism should therefore be evaluated against both a strong deterministic translator and a capacity-matched deterministic refiner, rather than only against unconditional or augmentation-oriented generators. Waveform fidelity in paired cross-location reconstruction should consequently be assessed directly rather than inferred from downstream HAR performance.

\subsection{Diffusion-Based Generative Modelling and Research Gap}

Diffusion models recover structured data from progressively corrupted states and have been applied to time-series generation, forecasting and imputation \cite{lin2024diffusion}. In wearable sensing, Oppel and Munz apply diffusion to multivariate IMU generation for HAR augmentation under limited data \cite{oppel2025diffusion}. Diffusion-based imputation further shows that observed temporal and feature context can constrain the denoising process \cite{islam2025selfattention,xiao2025boundary}. These studies establish the feasibility of conditional diffusion for inertial and temporal data, but they do not address paired body-location translation anchored simultaneously by an observed source signal and a deterministic target estimate.

Flow Matching provides a related generative alternative based on continuous vector-field transport \cite{lipman2023flow,tong2024improving}. For the present task, the relevant question is not whether one generative family is universally superior, but whether generative residual refinement improves reconstruction beyond strong deterministic and capacity-matched alternatives under identical held-out pairs.

The remaining gap is therefore a controlled cross-location formulation in which a deterministic translator captures the predictable target component and a diffusion model learns only the unresolved residual, with reconstruction quality, inference cost, realistic activity conditioning and true external zero-shot transfer evaluated explicitly. DiffusionRef-IMU is designed for this setting.

\begin{table}[!htbp]
\centering
\caption{Representative recent methods related to cross-location wearable sensing, virtual sensor generation, and conditional time-series modelling. The comparison is methodological and protocol-aware and does not imply direct quantitative comparability across datasets or tasks.}
\label{tab:related}
\scriptsize
\setlength{\tabcolsep}{3pt}
\begin{tabularx}{\linewidth}{@{}p{1.9cm}p{2.0cm}X X X X@{}}
\toprule
Method & Primary task & Core modelling strategy & Source-to-target conditioning / translation & Evaluation protocol & Relevance / distinction \\
\midrule
IMUCoCo \cite{zhou2025imucoco} & Flexible-placement pose/HAR & Coordinate-conditioned shared representation from variable on-body IMUs & Placement coordinates condition representation; no paired target-waveform synthesis & Pose estimation and HAR across flexible placements & Strong placement-robust representation model, but not explicit source-to-target signal translation \\
SensorGAN \cite{hussein2024sensorgan} & Missing wearable-sensor recovery & Conditional GAN reconstructing an unavailable sensor stream & Concurrent observed sensors $\rightarrow$ missing stream & Recovery followed by HAR on public wearable datasets & Explicit sensor reconstruction, but assumes multi-sensor context rather than held activity--location translation \\
SynHAR \cite{uhlenberg2024synhar} & Synthetic inertial augmentation & Personalised biomechanics and human-surface simulation & Motion/body model $\rightarrow$ virtual IMU; not paired observed-source conditioning & Synthetic/real fusion and HAR transfer & Placement-aware virtual IMU synthesis, but simulation/augmentation rather than learned sensor-to-sensor translation \\
Oppel \& Munz \cite{oppel2025diffusion} & Inertial time-series generation & Denoising diffusion model for multivariate IMU synthesis & Distribution/activity-conditioned generation; no paired source-location target & Synthetic generation plus downstream HAR utility & Recent inertial diffusion generation; different path and reconstruction objective \\
Tong et al. \cite{tong2024improving} & Conditional generative transport & Generalised conditional flow matching with flexible source distributions and minibatch OT & Conditioned continuous transport; not wearable-placement specific & Generative-modelling benchmarks & Closest flow formulation, but not Direct-guided cross-location IMU reconstruction \\
\bottomrule
\end{tabularx}
\end{table}

% ==================================================================
\section{Methodology}\label{sec:method}
% ==================================================================

\begin{figure}[pos=htbp]
\centering
\IfFileExists{Fig_1_DiffusionRef_IMU_Pipeline.png}{%
    \includegraphics[width=0.94\linewidth]
    {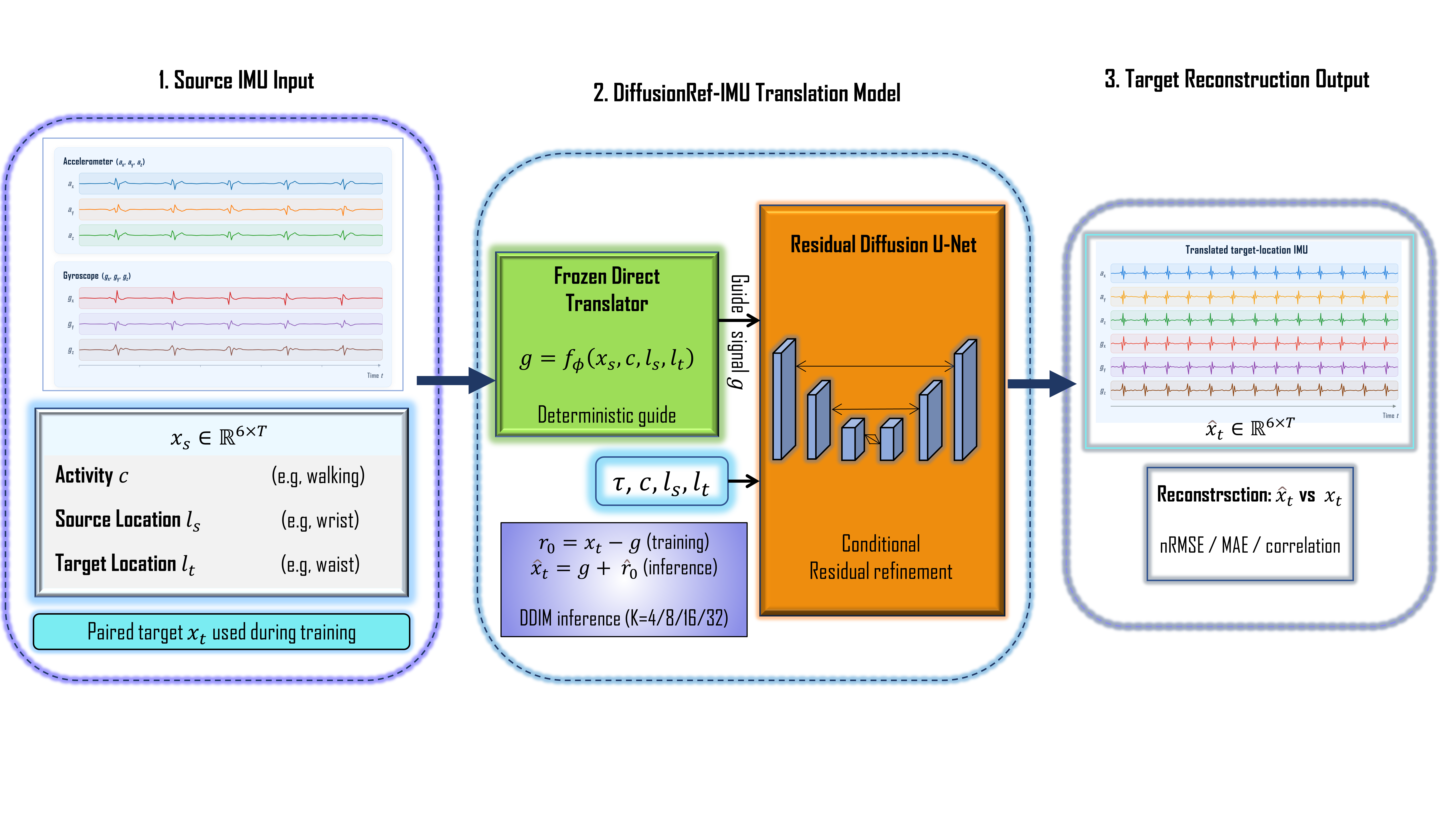}%
}{%
    \fbox{%
        \parbox[c][3.5cm][c]{0.90\linewidth}{%
            \centering
            \textbf{Figure 1 placeholder}\\[2mm]
            DiffusionRef-IMU pipeline:
            source IMU $\rightarrow$ Direct guide
            $\rightarrow$ residual diffusion
            $\rightarrow$ deterministic DDIM
            $\rightarrow$ target reconstruction.
        }%
    }%
}
\caption{Overview of DiffusionRef-IMU. A deterministic Direct translator first
produces a target-location guide. DiffusionRef-IMU models the unresolved
target residual while retaining the source signal and Direct guide as
persistent conditioning, and deterministic DDIM produces the final
target-location reconstruction.}
\label{fig:pipeline}
\end{figure}

This section defines DiffusionRef-IMU and the complete evaluation protocol. Figure~\ref{fig:pipeline} summarises the framework. A deterministic Direct translator first estimates the target-location inertial signal from an observed source-location window, and DiffusionRef-IMU then models only the residual between this guide and the paired target through conditional residual diffusion. The following subsections specify the diffusion objective and deterministic DDIM reconstruction, the main composition-held-out protocol, comparison models, source-only activity conditioning, true zero-shot external evaluation, and the statistical and implementation procedures.

% ------------------------------------------------------------------
\subsection{Problem Formulation and System Overview}
% ------------------------------------------------------------------

Let
$x_s\in\mathbb{R}^{6\times T}$ and
$x_t\in\mathbb{R}^{6\times T}$
denote synchronised accelerometer--gyroscope windows recorded at source and
target body locations $\ell_s$ and $\ell_t$, respectively. Each window contains
three accelerometer and three gyroscope channels, and $c$ denotes the
activity/context identifier. Cross-location IMU translation seeks to recover
the target waveform for the same activity and time interval from an observation
acquired at another body location.

A deterministic translator $f_{\phi}$ first generates the target-location
guide

\begin{equation}
g =
f_{\phi}
\left(
x_s,c,\ell_s,\ell_t
\right),
\label{eq:direct-guide}
\end{equation}

where $g\in\mathbb{R}^{6\times T}$. The remaining target discrepancy is

\begin{equation}
r_0 = x_t-g.
\label{eq:residual-target}
\end{equation}

DiffusionRef-IMU therefore does not synthesise the full target signal from an
independent prior. The deterministic network first captures the predictable
source-to-target component, and the diffusion model operates only on the
unresolved residual. After residual reconstruction, the translated target is

\begin{equation}
\widehat{x}_t = g+\widehat{r}_0.
\label{eq:final-reconstruction}
\end{equation}

The raw source window and Direct guide remain available to the residual network
at every diffusion step, with activity and source/target-location identities
providing additional conditioning. Ground-truth activity is used for the
controlled primary comparison; a source-only activity predictor is evaluated
separately to remove this requirement during deployment.

% ------------------------------------------------------------------
\subsection{Deterministic Guide and Guide-Anchored Residual Diffusion}
% ------------------------------------------------------------------

The Direct translator is a one-dimensional FiLM-conditioned U-Net that maps the
source waveform to a six-channel target estimate. Its conditioning combines a
128-dimensional sinusoidal temporal representation, a 64-dimensional activity
embedding, and 32-dimensional embeddings for the source and target locations.
These representations are projected into a 256-dimensional conditioning vector.
The temporal backbone uses channel widths 64, 128, 256 and 256 with
GroupNorm, SiLU activation and FiLM-modulated residual blocks. The archived EMA
Direct checkpoint associated with each fold is frozen before DiffusionRef-IMU
training and receives no gradient from the residual model.

DiffusionRef-IMU uses the same residual U-Net parameterisation as the
capacity-controlled residual experiments. Its input contains 18 channels,

\begin{equation}
\left[
r_t,\;x_s,\;g
\right],
\label{eq:diff-input}
\end{equation}

comprising the six-channel noisy residual state, the six-channel raw source
window and the six-channel Direct guide. The network retains base width 64,
conditioning dimension 256 and dropout 0.05, with 5,991,046 trainable
parameters. Its output is a six-channel $v$-prediction, and the final
convolution is zero-initialised.

A cosine diffusion schedule with
$T_d=1000$ discrete diffusion states defines the cumulative coefficients
$\bar{\alpha}_t$. During training, an integer diffusion index
$t\in\{0,\ldots,T_d-1\}$ is sampled uniformly, and Gaussian noise
$\epsilon\sim\mathcal{N}(0,I)$ is added to the residual:

\begin{equation}
r_t
=
\sqrt{\bar{\alpha}_t}\,r_0
+
\sqrt{1-\bar{\alpha}_t}\,\epsilon.
\label{eq:forward-diffusion}
\end{equation}

The implemented $v$-prediction target is

\begin{equation}
v_t^{\star}
=
\sqrt{\bar{\alpha}_t}\,\epsilon
-
\sqrt{1-\bar{\alpha}_t}\,r_0.
\label{eq:v-target}
\end{equation}

Using the normalised diffusion index
$\tau=t/(T_d-1)$, the model predicts

\begin{equation}
v_{\theta}
=
v_{\theta}
\left(
r_t,\tau,c,\ell_s,\ell_t,x_s,g
\right).
\label{eq:v-prediction}
\end{equation}

Given $v_{\theta}$, the corresponding clean residual estimate and predicted
noise are recovered as

\begin{align}
\widehat r_0
&=
\sqrt{\bar{\alpha}_t}\,r_t
-
\sqrt{1-\bar{\alpha}_t}\,
v_{\theta},
\label{eq:r0-recovery}
\\
\widehat\epsilon
&=
\sqrt{1-\bar{\alpha}_t}\,r_t
+
\sqrt{\bar{\alpha}_t}\,
v_{\theta}.
\label{eq:eps-recovery}
\end{align}

The implied target estimate is therefore
$\widetilde{x}_t=g+\widehat r_0$. This formulation centres the diffusion process
on the correction required beyond the frozen deterministic guide rather than
requiring the generative model to reconstruct the complete waveform
independently.

% ------------------------------------------------------------------
\subsection{Training Objective and Deterministic DDIM Inference}
% ------------------------------------------------------------------

The DiffusionRef-IMU objective combines the $v$-prediction error with
three reconstruction-oriented auxiliary terms:

\begin{equation}
\begin{split}
\mathcal{L}
={}&
\mathcal{L}_{v}
+
1.5\,\mathcal{L}_{\mathrm{MSE}}
+
0.05\,\mathcal{L}_{\mathrm{L1}}
+
0.02\,\mathcal{L}_{\Delta},
\end{split}
\label{eq:diffusion-loss}
\end{equation}

where

\begin{equation}
\mathcal{L}_{v}
=
\left\|
v_{\theta}-v_t^{\star}
\right\|_2^2,
\end{equation}

$\mathcal{L}_{\mathrm{MSE}}$ and
$\mathcal{L}_{\mathrm{L1}}$ compare
$\widetilde{x}_t$ with the paired target $x_t$, while
$\mathcal{L}_{\Delta}$ is the mean-squared error between their first temporal
differences. Unlike the preceding flow formulation, the final diffusion
objective includes neither a spectral term nor an anti-drift rollout.

Optimisation uses AdamW with learning rate $10^{-4}$ and weight decay
$10^{-4}$, with gradients clipped to unit norm. Training uses FP16 automatic
mixed precision, EMA decay 0.9998, and a cosine learning-rate schedule ending
at 5\% of the initial rate. Each fold is trained for exactly 50,000 optimiser
steps. A micro-batch size of 128 with two-step gradient accumulation gives an
effective batch size of 256. Validation is performed periodically for
diagnostic monitoring, but the primary reconstruction model is always the fixed
final 50k EMA checkpoint rather than a validation-selected diffusion
checkpoint.

Primary reconstruction uses deterministic DDIM with $\eta=0$. The paired
reconstruction endpoint does not initialise the residual from a random Gaussian
sample. Instead, the terminal residual state is set to zero, yielding a
deterministic mean-prior trajectory. A reproducible fixed-noise trajectory is
retained only as a generative diagnostic and is not used as the primary
reconstruction endpoint.

For consecutive inference indices $t$ and $t'$, the model first recovers
$\widehat r_0$ and $\widehat\epsilon$ from
Eqs.~\eqref{eq:r0-recovery}--\eqref{eq:eps-recovery}. The deterministic DDIM
update is then

\begin{equation}
r_{t'}
=
\sqrt{\bar{\alpha}_{t'}}\,
\widehat r_0
+
\sqrt{1-\bar{\alpha}_{t'}}\,
\widehat\epsilon .
\label{eq:ddim-update}
\end{equation}

Four inference budgets are evaluated: 4, 8, 16 and 32 DDIM steps. The fixed
final EMA model with 16-step DDIM is the primary accuracy endpoint, while
four-step DDIM is evaluated as the reduced-computation operating point. The
final target reconstruction is obtained from
Eq.~\eqref{eq:final-reconstruction} and clipped to the implementation range
$[-1.2,1.2]$.

\begin{algorithm}[H]
\caption{DiffusionRef-IMU deterministic residual reconstruction}
\label{alg:diffusionref}
\begin{algorithmic}[1]
\Require frozen Direct model $f_{\phi}$; trained EMA diffusion model
$v_{\theta}$; source window $x_s$; activity $c$; locations
$\ell_s,\ell_t$; cumulative schedule $\{\bar{\alpha}_t\}$; DDIM budget $K$
\State $g\gets f_{\phi}(x_s,c,\ell_s,\ell_t)$
\State Initialise residual state $r\gets 0$
\State Construct $K$ descending diffusion indices from $T_d-1$ to $0$
\For{each index $t$ in the DDIM schedule}
    \State $\tau\gets t/(T_d-1)$
    \State $v\gets
    v_{\theta}(r,\tau,c,\ell_s,\ell_t,x_s,g)$
    \State Recover $(\widehat r_0,\widehat\epsilon)$ using
    Eqs.~\eqref{eq:r0-recovery}--\eqref{eq:eps-recovery}
    \If{$t$ is the final scheduled index}
        \State $r\gets\widehat r_0$
    \Else
        \State Update $r$ using Eq.~\eqref{eq:ddim-update}
    \EndIf
\EndFor
\State $\widehat{x}_t\gets
\operatorname{clip}(g+r,-1.2,1.2)$
\State \Return $\widehat{x}_t$
\end{algorithmic}
\end{algorithm}

% ------------------------------------------------------------------
\subsection{Datasets, Preprocessing, and Activity--Location Holdout Protocol}
% ------------------------------------------------------------------

The main experiments use the Comprehensive IMU Dataset
\cite{feng2026comprehensive}, which comprises 30 participants, 12 activities and
17 synchronised body locations recorded at 60~Hz. Each location provides
three-axis acceleration and three-axis angular velocity. Signals are segmented
into two-second windows ($T=120$) with 50\% overlap. A location is eligible for
pairing when at least 98\% of its samples within the window are valid.

Normalisation statistics are estimated only from training subjects and
separately for each body location and channel. The 0.5th and 99.5th training
percentiles define a robust range that is mapped affinely to $[-1,1]$ and
clipped to this interval.

Five subject-disjoint folds are used. Each fold contains 20 training,
four validation and six test participants. The same fixed activity--location
composition map is withheld in every subject split:
Lying--Head, Sitting--LeftShoulder, Standing--LowerBack,
SlowWalking--RightUpperArm, ModerateWalking--LeftWrist,
BriskWalking--LeftShank, AscendingStairs--RightFoot,
DescendingStairs--RightForeArm, Cycling--RightThigh,
Running--LeftThigh, Jumping--LeftUpperArm and Rowing--RightShoulder.
The corresponding activity and anatomical location remain represented elsewhere
in training; only the specified composition is excluded.

During translator training, a withheld activity--location cell is excluded from
both source and target roles. Training source--target pairs change
deterministically across epochs, whereas validation pairing remains fixed. Test
evaluation uses a stored manifest specifying subject, activity, window, source
location and held target location. Direct, deterministic residual, flow and
diffusion models are therefore evaluated on identical test windows.

The primary DiffusionRef checkpoint is the fixed final 50k EMA state and is not
selected from the held-composition test set. The frozen Direct guide is inherited
from the archived deterministic baseline and retains its original
validation-based checkpoint-selection procedure, which used held-composition
validation subjects. Test participants remain disjoint; however, the complete
two-stage pipeline is therefore not described as composition-blind at every
model-selection stage.

\begin{table}[!htbp]
\centering
\caption{Datasets and evaluation protocols used in the study.}
\label{tab:datasets}
\scriptsize
\setlength{\tabcolsep}{3pt}
\begin{tabularx}{\linewidth}{
@{}p{1.8cm}p{1.45cm}p{1.4cm}p{1.25cm}p{1.5cm}X@{}}
\toprule
Dataset &
Subjects / records &
Activities &
Locations &
Native sampling &
Evaluation role \\
\midrule
Comprehensive IMU
\cite{feng2026comprehensive}
&
30 subjects
&
12
&
17
&
60 Hz
&
Five subject-disjoint activity--location composition-held-out folds;
120-sample windows with 50\% overlap.
\\

Grouvel2025
\cite{grouvel2025inertial}
&
59 subjects / 881 records
&
15 available
&
9
&
128 Hz
&
True zero-shot transfer using shared Walk semantics; two-second windows
resampled to the main 60-Hz representation.
\\

Salda\~na2025
\cite{rivas2025multisensor}
&
64 retained subjects
&
6 available
&
3
&
50 Hz
&
True zero-shot transfer using shared Walk and Bike semantics; two-second
windows resampled to the main 60-Hz representation.
\\
\bottomrule
\end{tabularx}
\end{table}

% ------------------------------------------------------------------
\subsection{Comparative Baselines and Source-Only Activity Conditioning}
% ------------------------------------------------------------------

DiffusionRef-IMU is the only proposed method. Three previously trained
translation systems are retained strictly as controlled reconstruction
comparators. Direct is the deterministic first-stage translator and also
provides the frozen guide used by DiffusionRef-IMU. DetRef applies a single
deterministic residual correction after the same Direct guide using the same
5,991,046-parameter residual U-Net parameterisation, but without a diffusion
process. CompFlow uses the same source/guide conditioning in a conditional
residual flow formulation with deterministic Heun integration. Together, these
controls separate improvements attributable to a second high-capacity residual
network from those associated with the diffusion objective itself.

\begin{table}[!htbp]
\centering
\caption{Translation systems retained for controlled reconstruction comparison.}
\label{tab:variants}
\small
\begin{tabularx}{\linewidth}{@{}p{2.2cm}p{2.1cm}p{2.3cm}X@{}}
\toprule
Variant &
Direct guide &
Second-stage model &
Evaluation role \\
\midrule
Direct
&
--
&
None
&
Strong deterministic reference and first-stage guide.
\\

DetRef
&
Frozen
&
One-shot deterministic residual U-Net
&
Capacity-matched deterministic residual control.
\\

CompFlow
&
Frozen
&
Conditional residual flow with Heun integration
&
Generative residual-transport comparator.
\\

\textbf{DiffusionRef-IMU}
&
Frozen
&
Guide-anchored residual diffusion
&
Proposed method.
\\
\bottomrule
\end{tabularx}
\end{table}

For the bounded main protocol, reconstruction error is measured by per-window
normalised RMSE,

\begin{equation}
\mathrm{nRMSE}
=
\frac{
\sqrt{
\frac{1}{6T}
\left\|
\widehat{x}_t-x_t
\right\|_F^2
}
}{2},
\label{eq:nrmse}
\end{equation}

where the denominator is the width of the normalised
$[-1,1]$ target range. Fold scores are obtained by averaging window-level
nRMSE values.

Because activity identity is part of the translation conditioning, a second
experiment replaces the oracle activity with a prediction derived solely from
the observed source IMU. The source-only classifier consists of a
six-to-64-channel Conv1d stem with kernel size seven and stride two, followed
by three temporal convolutional blocks with output widths 64, 128 and 256.
Global temporal mean pooling is concatenated with a 32-dimensional
source-location embedding and passed through a 256-unit fully connected layer
with dropout 0.2 before 12-way activity classification.

The activity predictor is trained only on source-side observations from the
training subjects using cross-entropy, AdamW with learning rate
$10^{-3}$ and weight decay $10^{-4}$, batch size 256 and FP16 mixed precision.
Training is limited to 50 epochs and stops after ten validation epochs without
improvement in macro-F1. The checkpoint with the best validation macro-F1 is
used. Translation is then evaluated on the same test manifest twice: once with
the recorded activity identity and once with the source-only predicted
activity. No target-location measurement is provided to the activity
predictor.

% ------------------------------------------------------------------
\subsection{True Zero-Shot External Evaluation}
% ------------------------------------------------------------------

External evaluation tests the five frozen main-dataset models without external
training, fine-tuning, checkpoint selection or normalisation fitting. External
signals are converted to two-second windows and resampled to 60~Hz, producing
the same 120-sample temporal representation used by the main models. Each
external window is then normalised using only the frozen location-specific
percentile statistics associated with the corresponding main-data fold. No
external statistics are estimated.

Grouvel2025 \cite{grouvel2025inertial} provides nine locations:
HE, LA, LF, LT, RA, RF, RT, SA and TR. These are mapped to
Head, LeftUpperArm, LeftForeArm, LeftThigh, RightUpperArm, RightForeArm,
RightThigh, LowerBack and UpperBack, respectively. Salda\~na2025
\cite{rivas2025multisensor} provides chest, left-knee and right-hand sensors,
mapped respectively to UpperBack, LeftShank and RightWrist.

Only activity semantics represented in the main vocabulary are used.
Grouvel2025 Walk is mapped to ModerateWalking, while Salda\~na2025 Walk and Bike
are mapped to ModerateWalking and Cycling, respectively. The primary zero-shot
experiment therefore evaluates sensor-domain and placement transfer without
introducing an unsupported activity class.

Native Grouvel windows contain 256 samples at 128~Hz with 50\% overlap, whereas
Salda\~na windows contain 100 samples at 50~Hz with 50\% overlap. Polyphase
resampling converts each selected segment to 120 samples. For each dataset,
ordered distinct source--target location pairs are assigned deterministically
across candidate windows. When the number of eligible windows exceeds the
evaluation limit, at most 4096 windows are selected using a fixed random seed.

Each external window is evaluated independently by all five frozen main-fold
models. Direct, DetRef, CompFlow and DiffusionRef therefore receive identical
external source/target pairs and identical fold-specific frozen normalisation
statistics. The primary DiffusionRef zero-shot endpoint uses the same
zero-prior deterministic DDIM reconstruction defined in
Section~\ref{sec:method}; no external diffusion update is performed.

The shared external activity labels are used for the primary controlled
zero-shot reconstruction because the external activity vocabularies differ
from the 12-class main-data vocabulary. Applying the main source-only activity
predictor externally is retained only as a secondary sensitivity analysis and
does not redefine the primary external endpoint.

% ------------------------------------------------------------------
\subsection{Statistical Analysis, Implementation, and Reproducibility}
% ------------------------------------------------------------------

For the main reconstruction experiment, the paired statistical unit is the
fold ($n=5$). For each comparison, fold-wise differences are oriented so that
positive values indicate improvement by the designated method. The analysis
reports the mean and median paired difference, sample standard deviation,
directional fold counts and a 95\% Student-$t$ confidence interval for the mean
paired difference. A two-sided exact sign-flip test enumerates all possible
sign assignments of the five paired differences. Standardised paired effects
are also expressed using Cohen's $d_z$ and its small-sample
bias-corrected Hedges $g_z$. Because only five independent folds are available,
exact-test resolution is inherently coarse; statistical interpretation
therefore considers direction, magnitude and interval estimates jointly rather
than relying on a significance threshold alone.

The zero-shot external analysis applies the same five frozen main-fold models
to common external windows. External results are therefore summarised
descriptively across the five frozen models rather than treated as five
independent external datasets.

All models are implemented in PyTorch, and training is executed on NVIDIA Tesla T4
GPUs using FP16 automatic mixed precision. Independent folds are assigned to
separate GPUs, with no gradient communication across folds. Fold-specific
random seeds derive from a common base seed of 42. Training checkpoints preserve
the network, EMA, optimiser, learning-rate scheduler, gradient scaler, global
step, epoch/batch position and random-number-generator states, allowing
interrupted execution to resume without intentionally altering the optimisation
schedule.

DiffusionRef inference cost is measured separately in FP32 with batch size one
on a Tesla T4. Each configuration receives 20 warm-up executions followed by
100 CUDA-event-timed repetitions. Median latency and the first and third
quartiles are retained. The number of refiner network evaluations equals the
number of DDIM steps; the complete pipeline also includes one Direct
forward pass.

The final experimental record stores fold definitions, test manifests,
normalisation statistics, model configurations, fixed-final EMA checkpoints
and fold-level evaluation outputs. Main reconstruction results are therefore
generated from locked manifests and saved predictions rather than from manually
reconstructed experimental settings.

These procedures define the basis for the reconstruction,
activity-conditioning, computational-efficiency and true zero-shot analyses
reported in Section~\ref{sec:results}.

% ==================================================================
\section{Results}\label{sec:results}
% ==================================================================

Results are reported in four stages. The primary five-fold cross-location
reconstruction experiment first compares DiffusionRef-IMU with Direct, DetRef
and CompFlow under the common held-composition protocol. The DDIM step ablation
is then examined jointly with measured T4 inference cost, followed by a
comparison of oracle and source-only predicted activity conditioning. Finally,
the frozen main-data models are evaluated under true zero-shot transfer to
Grouvel2025 and Salda\~na2025.

% ------------------------------------------------------------------
\subsection{Main Cross-Location Reconstruction}
% ------------------------------------------------------------------

Table~\ref{tab:main-recon} reports the primary reconstruction results.
The deterministic Direct translator achieved a mean nRMSE of
$0.090262\pm0.004937$, which the capacity-matched deterministic residual
refiner reduced to $0.088847\pm0.004969$. CompFlow achieved
$0.088366\pm0.005023$ with two-step Heun integration and
$0.088417\pm0.005049$ with the original eight-step configuration.

\begin{figure}[pos=htbp]
\centering
\IfFileExists{Fig_2_Reconstruction_Performance.png}{%
    \includegraphics[width=0.95\linewidth]
    {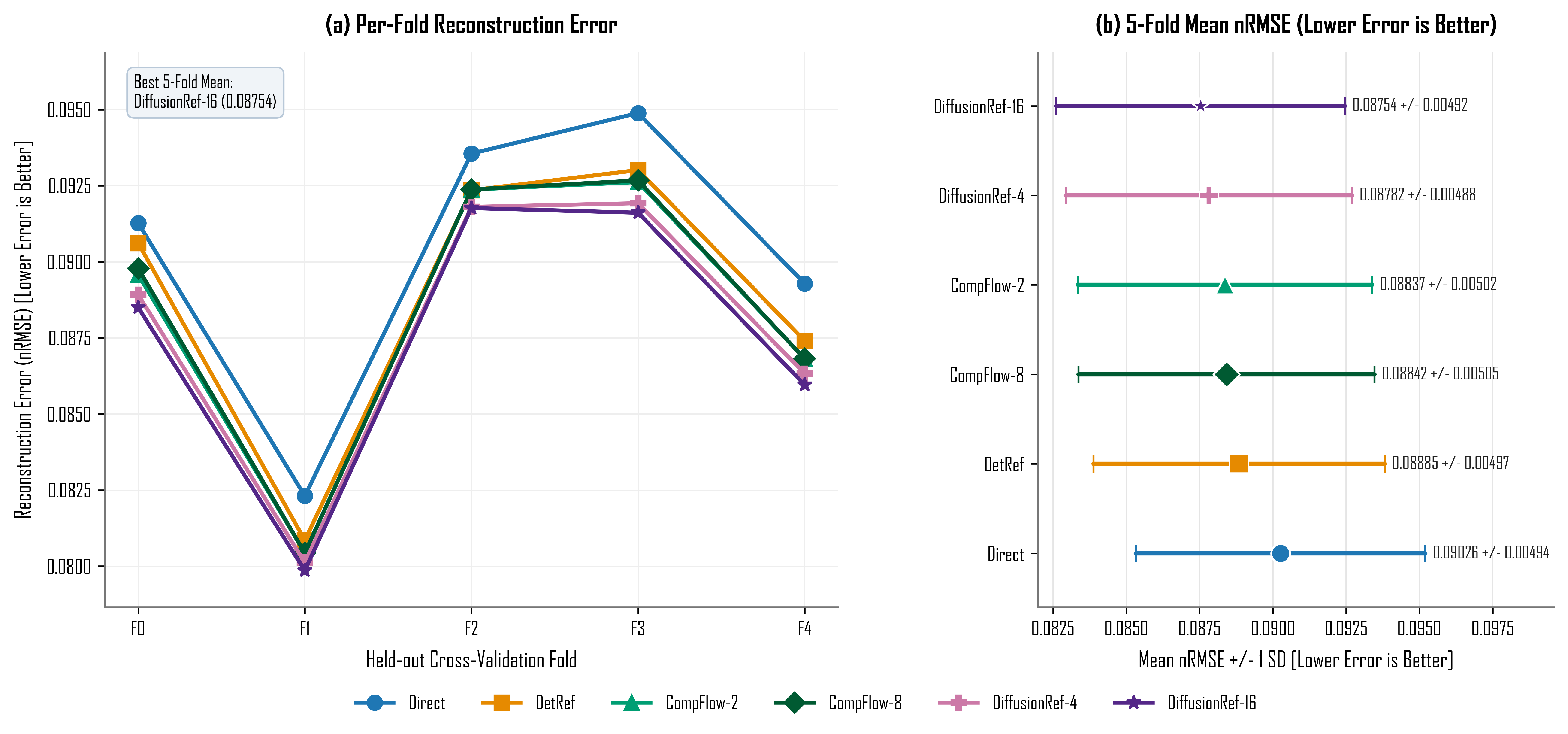}%
}{%
    \fbox{%
        \parbox[c][4cm][c]{0.90\linewidth}{%
        \centering
        Upload \texttt{Fig\_2\_Reconstruction\_Performance.png}
        }%
    }%
}
\caption{Five-fold cross-location reconstruction performance. The left panel
shows fold-wise nRMSE, while the right panel summarises mean and sample
standard deviation across folds. Lower values indicate better reconstruction.}
\label{fig:main-recon}
\end{figure}

DiffusionRef-IMU produced lower error than Direct, DetRef and eight-step
CompFlow in every fold. Four-step DDIM achieved
$0.087820\pm0.004885$, while the primary 16-step configuration achieved
$0.087538\pm0.004922$. Relative to Direct, DiffusionRef-16 reduced mean nRMSE
by 0.002724, corresponding to a 3.02\% relative reduction. The corresponding
reductions were 0.001309 (1.47\%) relative to DetRef and 0.000879 (0.99\%)
relative to CompFlow-8. Figure~\ref{fig:main-recon} shows that these differences
are directionally consistent across the five subject-disjoint folds.

\begin{table}[!htbp]
\centering
\caption{Five-fold cross-location reconstruction performance
(nRMSE; lower is better). DiffusionRef-16 is the primary reconstruction
configuration.}
\label{tab:main-recon}
\scriptsize
\setlength{\tabcolsep}{3.0pt}
\begin{tabular}{@{}lcccccc@{}}
\toprule
Method &
F0 &
F1 &
F2 &
F3 &
F4 &
Mean $\pm$ SD \\
\midrule

Direct
& 0.091271
& 0.082307
& 0.093557
& 0.094887
& 0.089286
& $0.090262\pm0.004937$ \\

DetRef
& 0.090614
& 0.080853
& 0.092351
& 0.093014
& 0.087403
& $0.088847\pm0.004969$ \\

CompFlow-2
& 0.089580
& 0.080438
& 0.092375
& 0.092621
& 0.086815
& $0.088366\pm0.005023$ \\

CompFlow-8
& 0.089785
& 0.080433
& 0.092378
& 0.092673
& 0.086814
& $0.088417\pm0.005049$ \\

DiffusionRef-4
& 0.088920
& 0.080121
& 0.091804
& 0.091926
& 0.086327
& $0.087820\pm0.004885$ \\

\textbf{DiffusionRef-16}
& \textbf{0.088499}
& \textbf{0.079857}
& \textbf{0.091766}
& \textbf{0.091613}
& \textbf{0.085956}
& $\mathbf{0.087538\pm0.004922}$ \\

\bottomrule
\end{tabular}
\end{table}

The paired analysis yields the same directional conclusion, although inference
is limited by the five-fold sample size. DiffusionRef-16 improved over Direct
in all 5/5 folds, with a mean paired reduction of 0.002724
(95\% CI 0.001934--0.003513; Hedges $g_z=3.427$). Against DetRef, the mean
reduction was 0.001309 (95\% CI 0.000601--0.002016;
$g_z=1.837$), again with 5/5 favourable folds. Relative to CompFlow-8,
DiffusionRef-16 reduced nRMSE by 0.000879
(95\% CI 0.000505--0.001252; $g_z=2.339$), also in all five folds.
Each 5/5 directional comparison yields an exact two-sided sign-flip
$p$-value of 0.0625. This value is not interpreted as conventional statistical
significance because, with five paired folds, 0.0625 is the smallest attainable
two-sided value for a uniform directional outcome.

% ------------------------------------------------------------------
\subsection{Inference-Step Ablation and Computational Efficiency}
% ------------------------------------------------------------------

\begin{figure}[pos=htbp]
\centering
\IfFileExists{Fig_3_Efficiency_Quality_Tradeoff.png}{%
    \includegraphics[width=0.88\linewidth]
    {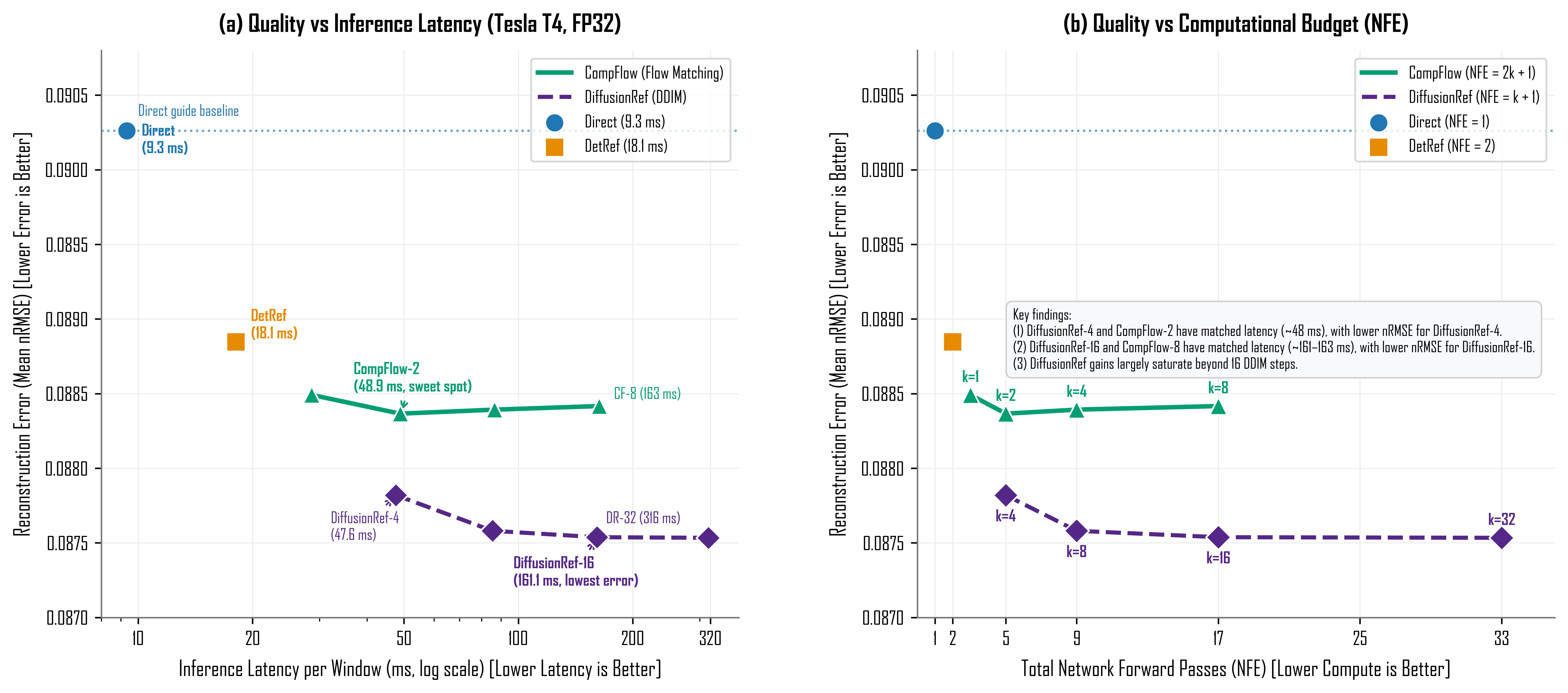}%
}{%
    \fbox{%
        \parbox[c][4cm][c]{0.84\linewidth}{%
        \centering
        Upload \texttt{Fig\_3\_Efficiency\_Quality\_Tradeoff.png}
        }%
    }%
}
\caption{Reconstruction quality versus median batch-1 FP32 inference latency
on an NVIDIA Tesla T4. CompFlow and DiffusionRef trajectories correspond to
increasing iterative-refinement budgets; lower and further left is preferable.}
\label{fig:efficiency}
\end{figure}

Figure~\ref{fig:efficiency} relates reconstruction quality to measured
batch-1 FP32 latency on a Tesla T4. Direct required 9.34~ms per window and
DetRef required 18.08~ms. For CompFlow, increasing Heun integration from one to
eight steps raised median latency from 28.59 to 163.13~ms. The lowest mean
CompFlow error occurred with two steps
($0.088366$ nRMSE, 48.89~ms), whereas the original eight-step configuration
required 163.13~ms for a mean nRMSE of 0.088417.

DiffusionRef showed a similar monotonic increase in computation but a lower
error frontier. Four-step DDIM achieved 0.087820 nRMSE at 47.62~ms, eight steps
achieved 0.087581 at 85.50~ms, and 16 steps achieved 0.087538 at
161.11~ms. Increasing inference to 32 steps reduced the mean only marginally
to 0.087534 while increasing latency to 316.15~ms.

The matched-compute comparisons are especially informative. CompFlow-2 and
DiffusionRef-4 both require four refiner evaluations plus one Direct forward
pass; DiffusionRef-4 was slightly faster (47.62 versus 48.89~ms) while
reducing mean nRMSE from 0.088366 to 0.087820. Likewise, CompFlow-8 and
DiffusionRef-16 each require 16 refiner evaluations plus one Direct forward
pass. DiffusionRef-16 required 161.11~ms versus 163.13~ms for CompFlow-8 while
lowering nRMSE from 0.088417 to 0.087538. The diffusion advantage therefore
remains when the iterative models are compared at approximately equal
network-forward counts and latency.

The 32-step result indicates saturation rather than a practically meaningful
additional gain: doubling the inference budget from 16 to 32 steps changes
mean nRMSE by only 0.000004. Accordingly, 16-step DDIM is retained as the
primary accuracy configuration, while four-step DDIM provides the more
practical quality--latency operating point. DiffusionRef-4 nevertheless
remains approximately 2.63$\times$ slower than the one-shot DetRef pipeline,
so the reconstruction gain carries additional computational cost.

% ------------------------------------------------------------------
\subsection{Robustness to Source-Only Activity Conditioning}
% ------------------------------------------------------------------

The source-only activity predictor achieved an overall classification accuracy
of 0.8478 and macro-F1 of 0.8473 across the five main folds. Table~
\ref{tab:activity-robustness} reports reconstruction when the recorded activity
identity is replaced by the classifier prediction at inference. All translation
variants show only a small increase in nRMSE.

\begin{figure}[pos=htbp]
\centering
\IfFileExists{Fig_4_Activity_Label_Robustness.png}{%
    \includegraphics[width=0.90\linewidth]
    {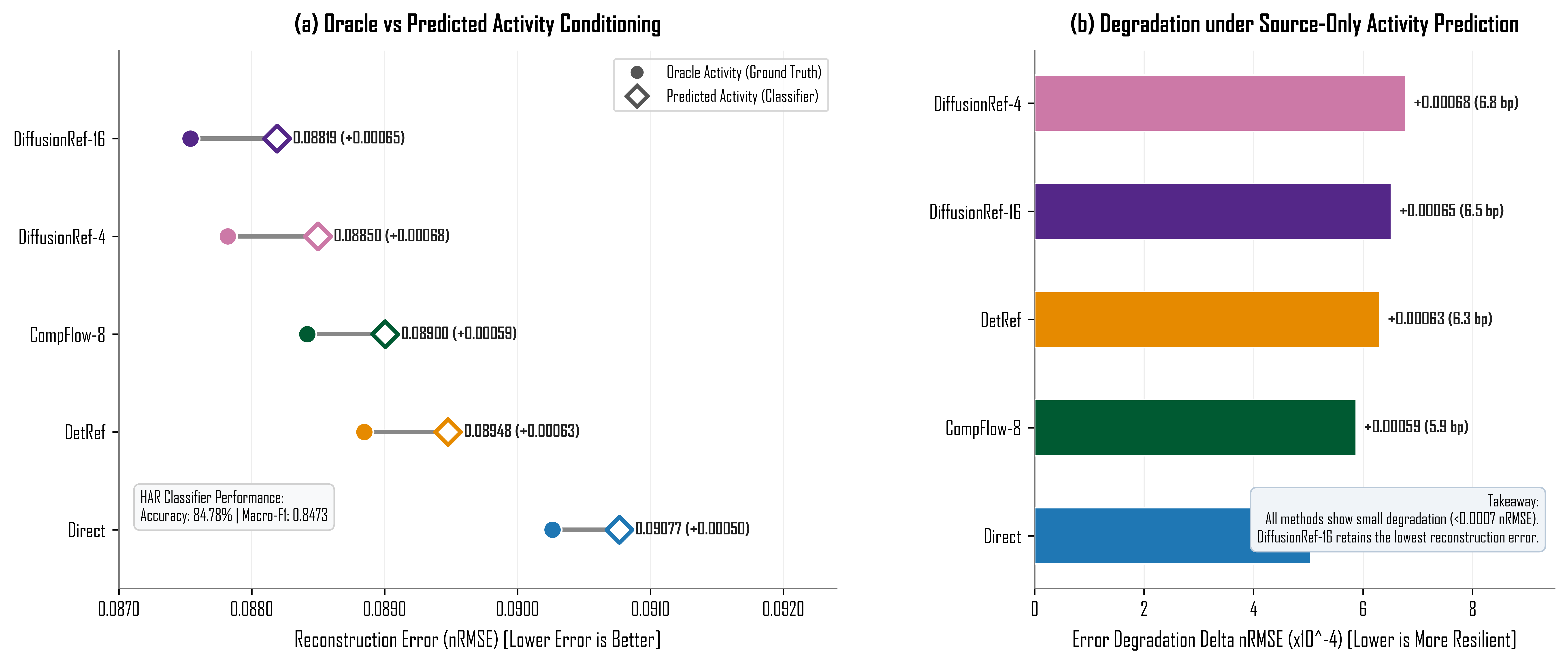}%
}{%
    \fbox{%
        \parbox[c][4cm][c]{0.86\linewidth}{%
        \centering
        Upload \texttt{Fig\_4\_Activity\_Label\_Robustness.png}
        }%
    }%
}
\caption{Reconstruction robustness when recorded activity labels are replaced
by predictions from the source-only activity classifier. The annotations show
the increase in mean nRMSE relative to oracle activity conditioning.}
\label{fig:activity-robustness}
\end{figure}

For DiffusionRef-16, mean nRMSE increased from 0.087538 with oracle activity
to 0.088190 with source-predicted activity, an absolute degradation of
0.000652. DiffusionRef-4 changed from 0.087820 to 0.088498
($+0.000678$). The corresponding changes were $+0.000504$ for Direct,
$+0.000630$ for DetRef and $+0.000587$ for CompFlow-8.
Figure~\ref{fig:activity-robustness} visualises these oracle-to-predicted
differences.

\begin{table}[!htbp]
\centering
\caption{Effect of replacing oracle activity conditioning with the source-only
activity prediction. Values are mean five-fold nRMSE; lower is better.}
\label{tab:activity-robustness}
\small
\setlength{\tabcolsep}{5pt}
\begin{tabular}{@{}lccc@{}}
\toprule
Method &
Oracle activity &
Predicted activity &
$\Delta$ nRMSE \\
\midrule

Direct
& 0.090262
& 0.090766
& +0.000504 \\

DetRef
& 0.088847
& 0.089477
& +0.000630 \\

CompFlow-8
& 0.088417
& 0.089004
& +0.000587 \\

DiffusionRef-4
& 0.087820
& 0.088498
& +0.000678 \\

\textbf{DiffusionRef-16}
& \textbf{0.087538}
& \textbf{0.088190}
& +0.000652 \\

\bottomrule
\end{tabular}

\par\vspace{2pt}
\footnotesize
Source-only activity predictor:
accuracy $=0.8478$, macro-F1 $=0.8473$.
\end{table}

This degradation remains small relative to the overall difference between
Direct and DiffusionRef-16. In particular, DiffusionRef-16 with
source-predicted activity remains numerically below the oracle-conditioned
Direct, DetRef and CompFlow-8 means. The experiment therefore removes the
requirement for a ground-truth activity label in the main deployment setting
while retaining most of the reconstruction improvement.

% ------------------------------------------------------------------
\subsection{True Zero-Shot External Transfer}
% ------------------------------------------------------------------

Table~\ref{tab:zero-shot} reports true zero-shot evaluation with all
main-data models, checkpoints and normalisation statistics kept frozen.
No external training, fine-tuning or normalisation fitting is performed.
The two external datasets yield different method orderings.

On Grouvel2025, Direct achieved
$0.405685\pm0.001760$ nRMSE. DiffusionRef-16 reduced the mean to
$0.405295\pm0.001872$ and was favourable relative to Direct in four of the
five frozen main-fold models. The absolute mean reduction was 0.000390,
equivalent to approximately 0.10\%. DiffusionRef-4 obtained 0.405847,
while DetRef and CompFlow-8 obtained 0.405730 and 0.405920, respectively.
The Grouvel result therefore provides a small favourable zero-shot effect for
the 16-step diffusion configuration.

The ordering differed on Salda\~na2025. Direct achieved the lowest mean nRMSE at
$0.380318\pm0.003221$. DiffusionRef-16 obtained
$0.381124\pm0.003127$, corresponding to an absolute degradation of 0.000806
(0.21\%) and improving over Direct in only one of the five frozen models.
DiffusionRef-4 obtained 0.381417, while DetRef and CompFlow-8 obtained
0.381120 and 0.381134. The primary internal reconstruction advantage therefore
does not transfer uniformly across external domains.

\begin{table}[!htbp]
\centering
\caption{True zero-shot external reconstruction. Values are mean nRMSE
$\pm$ sample SD across the five frozen main-fold models. No external training,
fine-tuning or normalisation fitting is used.}
\label{tab:zero-shot}
\small
\setlength{\tabcolsep}{4.2pt}
\begin{tabular}{@{}lcc@{}}
\toprule
Method &
Grouvel2025 &
Salda\~na2025 \\
\midrule

Direct
& $0.405685\pm0.001760$
& $\mathbf{0.380318\pm0.003221}$ \\

DetRef
& $0.405730\pm0.001858$
& $0.381120\pm0.003268$ \\

CompFlow-8
& $0.405920\pm0.001935$
& $0.381134\pm0.003289$ \\

DiffusionRef-4
& $0.405847\pm0.001799$
& $0.381417\pm0.003225$ \\

\textbf{DiffusionRef-16}
& $\mathbf{0.405295\pm0.001872}$
& $0.381124\pm0.003127$ \\

\bottomrule
\end{tabular}
\end{table}

The frozen main-data normalisation also reveals substantial external
distribution shift. Mean source/target clipping fractions were approximately
0.552/0.546 on Grouvel2025 and 0.482/0.480 on Salda\~na2025. Direct
transfer of the main 12-class source-activity predictor to the external
vocabularies produced essentially no semantic label agreement
(0 on Grouvel2025 and approximately 0.003 on Salda\~na2025). Under that
secondary sensitivity condition, DiffusionRef-16 nRMSE degraded to 0.410455
and 0.384407, respectively. These values are therefore not used as the
primary external endpoint; the controlled zero-shot comparison in
Table~\ref{tab:zero-shot} uses the shared activity-semantic mapping specified
in Section~\ref{sec:method}.

\begin{figure}[pos=htbp]
\centering
\IfFileExists{Fig_5_True_ZeroShot_External_Transfer.png}{%
    \includegraphics[width=0.90\linewidth]
    {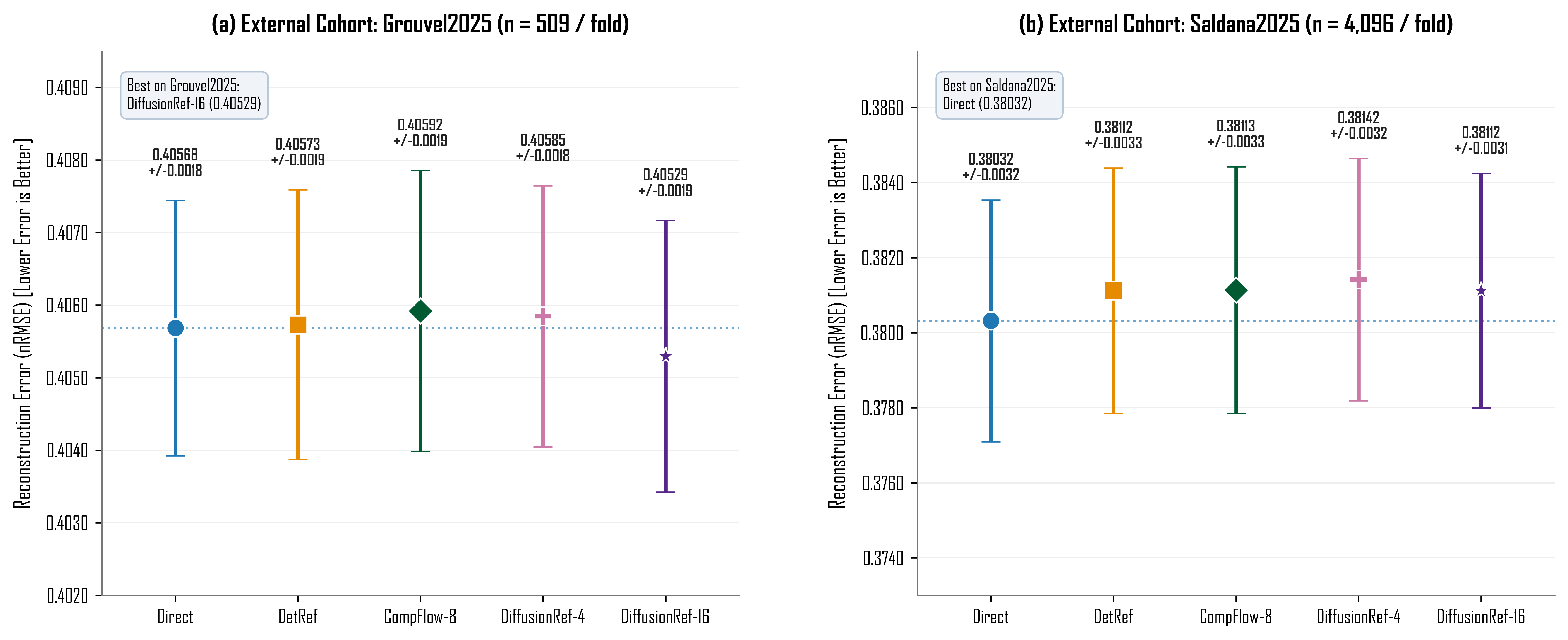}%
}{%
    \fbox{%
        \parbox[c][4cm][c]{0.86\linewidth}{%
        \centering
        Upload \texttt{Fig\_5\_True\_ZeroShot\_External\_Transfer.png}
        }%
    }%
}
\caption{True zero-shot reconstruction on Grouvel2025 and Salda\~na2025.
DiffusionRef-16 gives the lowest mean nRMSE on Grouvel2025, whereas Direct
remains strongest on Salda\~na2025.}
\label{fig:zero-shot}
\end{figure}

Figure~\ref{fig:zero-shot} summarises the primary external comparison.
DiffusionRef-16 is marginally favourable on Grouvel2025, whereas Direct
remains preferable on Salda\~na2025. The external evidence therefore supports
dataset-dependent rather than universal zero-shot transfer.

% ------------------------------------------------------------------
% Consolidated findings
% ------------------------------------------------------------------

Taken together, four findings define the experimental scope. First,
DiffusionRef-16 improves reconstruction over Direct, DetRef and CompFlow in all
five main folds, with a 3.02\% mean reduction relative to Direct. Second, the
improvement persists at matched iterative compute: four-step DiffusionRef
outperforms two-step CompFlow at approximately 48~ms, while 16-step
DiffusionRef outperforms eight-step CompFlow at approximately 162~ms; gains
saturate beyond 16 DDIM steps. Third, replacing oracle activity conditioning
with a source-only predictor introduces only a small reconstruction penalty on
the main protocol. Fourth, true zero-shot transfer is mixed: DiffusionRef-16 is
marginally favourable on Grouvel2025 but does not improve over Direct on
Salda\~na2025. These findings establish the empirical boundary of
DiffusionRef-IMU without implying universal cross-dataset superiority.

% ==================================================================
\section{Discussion}\label{sec:discussion}
% ==================================================================

\subsection{Principal Findings}

The central finding is that a strong deterministic cross-location estimate
provides an effective anchor for residual diffusion. DiffusionRef-IMU does not
generate the complete target waveform independently; Direct first captures the
predictable source-to-target component, and the diffusion model refines only
the remaining discrepancy. Across the five subject-disjoint,
activity--location composition-held-out folds, 16-step DiffusionRef achieved
the lowest mean reconstruction error and improved over Direct, DetRef and
eight-step CompFlow in every fold. The larger improvement over Direct relative
to the remaining differences among residual methods indicates that
deterministic guide construction accounts for an important part of the overall
translation performance.

The capacity-matched DetRef comparison is central to interpreting the remaining
gain. DetRef uses the same Direct guide and an equally parameterised residual
network but removes the diffusion process. DiffusionRef-16 reduced mean nRMSE
from 0.088847 for DetRef to 0.087538, with a favourable direction in all five
folds. This controlled comparison indicates that the improvement cannot be
explained solely by adding a second high-capacity refinement network. The
absolute difference, however, remains modest, and the exact two-sided sign-flip
test yields $p=0.0625$. With only five paired folds, this is the minimum
attainable two-sided value for a uniform 5/5 directional outcome. The evidence
therefore supports a consistent reconstruction advantage without constituting a
conventional $p<0.05$ significance claim.

The inference ablation clarifies the practical contribution. Four-step
DiffusionRef retained most of the reconstruction benefit at 47.62~ms per window
and outperformed two-step CompFlow at essentially matched latency and
network-forward count. Increasing DDIM inference to 16 steps provided the
primary accuracy endpoint, whereas 32 steps produced negligible additional
improvement despite approximately doubling latency. The advantage is therefore
better characterised as a quality--computation frontier than as unrestricted
benefit from additional diffusion steps. DiffusionRef-4 also remains slower
than the one-shot DetRef pipeline, so the increase in fidelity is obtained at a
non-zero computational cost.

Activity conditioning defines a second practical boundary. Replacing the
recorded activity label with a source-only predictor with macro-F1 0.8473
increased DiffusionRef-16 nRMSE only from 0.087538 to 0.088190. This result
suggests that the main reconstruction benefit does not depend critically on
access to ground-truth activity identity at deployment. Nevertheless, activity
prediction remains part of the current conditional formulation rather than
being eliminated entirely.

The external results impose the strongest constraint on generalisation claims.
Without external training, fine-tuning or normalisation fitting,
DiffusionRef-16 was marginally favourable on Grouvel2025 but did not improve
on Direct for Salda\~na2025. The internal five-fold advantage therefore does not
transfer uniformly across datasets. The evidence supports dataset-dependent
zero-shot transfer rather than universal cross-dataset superiority.

\subsection{Interpretation in Relation to Prior Work}

Wearable placement research commonly addresses changes in body location by
learning representations that remain useful for classification rather than by
reconstructing the waveform that another sensor would have measured.
Cross-body adaptation therefore addresses a different problem from paired
source-to-target translation \cite{hegde2023activity}. Likewise, virtual sensor
approaches such as WIMUSim and SynHAR synthesise inertial observations through
physics or biomechanical models \cite{oishi2025wimusim,uhlenberg2024synhar},
while SensorGAN recovers unavailable wearable streams from concurrently
observed sensors \cite{hussein2024sensorgan}. Diffusion-based IMU work has also
demonstrated the utility of generative models for synthetic time-series
production and augmentation \cite{oppel2025diffusion}. These methods establish
the broader value of virtual sensing, but their objectives and evaluation
protocols are not directly comparable with paired cross-location
reconstruction.

DiffusionRef-IMU occupies a narrower methodological position. The model is
conditioned on a specific observed source window and a deterministic estimate
of the paired target, and diffusion is applied only to the unresolved residual.
Its deterministic zero-prior DDIM inference should therefore be interpreted as
conditional residual reconstruction rather than conventional stochastic sample
generation. This design is consistent with the broader use of conditional
diffusion for structured time-series recovery
\cite{lin2024diffusion,islam2025selfattention,xiao2025boundary}, but adapts that
principle to anatomically conditioned source-to-target translation.

The deterministic and flow-based controls are important because an apparent
generative advantage can otherwise reflect architecture size or iterative
computation rather than the modelling objective. DetRef substantially improves
over Direct despite containing no generative process, confirming that residual
refinement itself is a strong component of the solution. The smaller remaining
DiffusionRef advantage therefore reflects the controlled difference between
deterministic and diffusion-based residual refinement under matched capacity.
The comparison with CompFlow adds a second control: at approximately matched
iterative inference budgets, DiffusionRef attains lower reconstruction error.
This result does not establish diffusion as universally superior to flow
matching, but it shows that the diffusion formulation provides the stronger
residual-refinement result under the present protocol.

The Grouvel--Salda\~na divergence should not be assigned a single causal
explanation. Differences in sensor layouts, activity structure, sampling
characteristics, anatomical correspondence and distribution shift may
contribute, but the present experiments do not isolate these factors. The
substantial clipping observed after applying frozen main-data normalisation
supports the presence of external distribution mismatch, although it does not
identify which source of mismatch drives the different method ordering.

\subsection{Implications for Cross-Location Wearable Sensing}

Several implications follow from these results. First, cross-location
generative translation should be evaluated against a strong deterministic
translator rather than identity or weak reconstruction baselines. Second, an
equally expressive deterministic residual model is required when the proposed
method adds a second high-capacity refinement network; otherwise, capacity and
modelling-objective effects cannot be separated. Third, iterative generative
methods should be compared at matched inference budgets. The near-equal-latency
DiffusionRef-4 versus CompFlow-2 and DiffusionRef-16 versus CompFlow-8
comparisons provide a clearer efficiency assessment than comparisons limited to
maximum-step configurations.

The source-only activity experiment also illustrates the importance of testing
the information actually available at deployment. Conditioning on oracle
activity can overstate practical accessibility when the label must first be
inferred from the observed sensor. Reporting both conditions separates the
behaviour of the translation model from the quality of its auxiliary context.

External zero-shot evaluation is equally important. Strong internal
cross-validation does not guarantee that frozen source-to-target mappings will
remain favourable under different sensors and acquisition distributions. The
different method orderings on Grouvel2025 and Salda\~na2025 therefore define a
meaningful deployment boundary rather than an inconsistency to be removed.
Transparent reporting of adverse external evidence, fixed final checkpoints
and historical model-selection choices provides a more reliable basis for
judging generalisation than reporting only favourable internal results.

Computational cost also remains part of the practical interpretation.
Four-step DDIM provides the most favourable operating point observed here,
retaining most of the accuracy gain while avoiding the substantially higher
cost of 16--32 steps. It should nevertheless not be described as
computationally equivalent to one-shot deterministic refinement. When latency
is more important than the remaining reconstruction gain, DetRef may still be
the more appropriate operating point.

\subsection{Limitations}

Several limitations bound the present evidence. The paired inferential unit is
the fold, and only five folds are available. Exact sign-flip tests therefore
have coarse resolution, and even a consistent 5/5 directional outcome cannot
produce a two-sided $p$-value below 0.0625. Fold direction, effect magnitude
and confidence intervals are consequently more informative than a binary
significance threshold.

The external evidence is limited to two datasets and is explicitly mixed.
DiffusionRef is marginally favourable on Grouvel2025 but not on
Salda\~na2025, so the method has not been shown to improve arbitrary external
sensor domains. External evaluation also requires anatomical and activity
semantic mapping into the main-data vocabulary. The large clipping fractions
under frozen main-data normalisation indicate substantial domain shift, while
the main source-activity classifier does not transfer cleanly to the external
activity vocabularies. These factors limit conclusions about fully autonomous
zero-shot deployment.

The DiffusionRef model uses fixed final 50k EMA checkpoints, but the frozen
Direct guides retain the historical validation-based checkpoint-selection
procedure described in Section~\ref{sec:method}. The full two-stage pipeline
therefore should not be described as composition-blind at every model-selection
stage. In addition, the revised study focuses on waveform reconstruction; a
DiffusionRef-specific downstream HAR augmentation experiment was not conducted,
and no downstream superiority claim is made. Finally, the evidence is restricted
to six-channel inertial signals, the studied activities, sensor locations,
participants and two-second windowing protocol and should not be extrapolated
directly to other wearable modalities or clinical applications.

\subsection{Future Work}

Future work should first adopt fully composition-blind selection for both the
Direct guide and residual model so that no held activity--location composition
contributes to model selection at any stage. Broader zero-shot evaluation on
additional independent datasets and sensor layouts is also required to
determine which forms of anatomical and distribution shift are compatible with
the learned translation.

More transferable conditioning could reduce dependence on dataset-specific
activity and location vocabularies. Anatomical descriptors, learned spatial
embeddings or activity-free conditioning are possible directions, while
domain-adaptive or test-time adaptation could be examined when fully frozen
zero-shot transfer is insufficient. A dedicated DiffusionRef downstream HAR
study would also establish whether the reconstruction advantage
translates into task-level utility.

Computational work should focus on the four-step operating point rather than on
further increasing the sampling trajectory. Distillation, consistency-style
training or adaptive DDIM step selection may preserve reconstruction quality
while narrowing the latency gap to deterministic residual refinement.
Uncertainty-aware residual generation could also be investigated where a single
deterministic reconstruction is insufficient to represent the possible
target-location signals.

% ==================================================================
\section{Conclusion}\label{sec:conclusion}
% ==================================================================

This study presented DiffusionRef-IMU, a guide-anchored residual diffusion
framework for cross-location inertial measurement unit translation. A
deterministic Direct network first produces a target-location estimate, after
which conditional diffusion models only the residual correction unresolved by
that guide. The formulation was evaluated under five subject-disjoint,
activity--location composition-held-out folds together with a
capacity-matched deterministic residual control, a conditional flow baseline,
source-only activity conditioning, inference-step ablation and true zero-shot
external evaluation.

DiffusionRef-16 achieved the lowest main reconstruction error, with mean nRMSE
0.087538 compared with 0.090262 for Direct, 0.088847 for DetRef and 0.088417
for eight-step CompFlow, and was favourable in all five folds against each
baseline. Four-step DDIM retained a mean nRMSE of 0.087820 at 47.62~ms,
providing the strongest observed quality--latency operating point. Replacing
oracle activity with a source-only predictor increased the 16-step result only
to 0.088190. External evidence was less uniform: DiffusionRef-16 was
marginally favourable on Grouvel2025, whereas Direct remained stronger on
Salda\~na2025.

These results support guide-anchored residual diffusion as an effective
controlled approach for improving cross-location reconstruction beyond
deterministic and flow-based residual refinement under the main protocol. They
do not support universal superiority or uniform zero-shot generalisation.
Broader independent datasets, stricter composition-blind model selection, more
transferable conditioning and lower-cost diffusion inference are required to
establish where the observed benefit persists across wearable sensing domains.

\section*{Acknowledgments}

The authors acknowledge American International University-Bangladesh (AIUB)
for its institutional support. The authors also acknowledge the Advanced
Machine Intelligence Research Lab (AMIR) for its research support and
facilities.

\section*{Conflicts of Interest}

The authors declare that they have no known competing financial interests or
personal relationships that could have appeared to influence the work reported
in this paper.

\section*{Data Availability Statement}

All datasets used in this study are publicly available from their official
repositories.

The Comprehensive IMU Dataset used for the main experiments is publicly
available through Figshare at:
\url{https://doi.org/10.6084/m9.figshare.30234940}.

The inertial dataset of daily living tasks used for the Grouvel2025
external validation is publicly available through Zenodo at:
\url{https://doi.org/10.5281/zenodo.15081742}.

The multi-sensor human activity recognition dataset used for the
Salda\~na2025 external validation is publicly available through Zenodo at:
\url{https://doi.org/10.5281/zenodo.15830858}.

No privately collected participant data were introduced in this study.
The subject-disjoint fold definitions, held activity--location compositions,
preprocessing procedures, model configurations, training protocols, and
evaluation methodology used for the reported experiments are described in
the manuscript. Code and processed experimental artefacts associated with
DiffusionRef-IMU will be made publicly available through the project repository upon publication.

\section*{CRediT authorship contribution statement}

\textbf{M.\,M. Golam Hafiz:}
Conceptualization, Methodology, Software, Data Curation, Investigation,
Formal Analysis, Validation, Visualization, Writing -- Original Draft,
Writing -- Review \& Editing.

\textbf{Safia Akter Simi:}
Conceptualization, Methodology, Software, Data Curation, Investigation,
Formal Analysis, Validation, Visualization, Writing -- Original Draft,
Writing -- Review \& Editing.

\textbf{Md. Rakibul Islam:}
Supervision, Project Administration, Methodology, Resources,
Funding Acquisition, Writing -- Review \& Editing.

\textbf{M.\,F. Mridha:}
Supervision, Project Administration, Methodology, Resources,
Writing -- Review \& Editing.

\section*{Declaration of Generative AI and AI-Assisted Technologies}

All experimental results, numerical values, statistical analyses, tables,
and figures reported in this study were produced directly from the authors'
implemented code, executed experiments, and verified experimental outputs.
No generative AI system was used to generate, modify, fabricate, or infer
experimental data, numerical results, statistical outcomes, plots, figures,
or scientific conclusions. No AI-based image-generation system was used to
create the manuscript figures or the methodological architecture diagram.

Generative AI tools were used only during manuscript preparation for
language refinement and organisational assistance. The authors independently
verified, revised, and take full responsibility for the final scientific
content of the manuscript.

\bibliographystyle{elsarticle-num-names}
\bibliography{DiffusionRef_IMU_References}

\end{document}